\documentclass{article}
\usepackage{arxiv}
\usepackage[utf8]{inputenc} 
\usepackage[T1]{fontenc}    
\usepackage{hyperref}       
\usepackage{url}            
\usepackage{booktabs}       
\usepackage{amsfonts}       
\usepackage{nicefrac}       
\usepackage{microtype}      
\usepackage{lipsum}
\usepackage{times}
\usepackage{helvet}
\usepackage{courier}  
\usepackage{graphicx}  
\usepackage{caption}
\usepackage{soul}
\usepackage{amsmath,bm}
\usepackage{amsfonts}
\usepackage{amssymb}
\usepackage{graphicx}
\usepackage[export]{adjustbox}
\usepackage{tabularx}
\usepackage{color}
\usepackage{multicol}
\usepackage[english]{babel}
\usepackage[utf8]{inputenc}
\usepackage{algorithm}
\usepackage[noend]{algpseudocode}

\usepackage[table]{xcolor}
\usepackage{tabularx}
\usepackage{enumitem}

\setlist{nolistsep}
\definecolor{green}{HTML}{66FF66}
\definecolor{myGreen}{HTML}{009900}

\usepackage{makecell, cellspace, caption}

\usepackage{longtable}
\newcommand{\mbfx}{\mathbf{x}}
\newcommand{\mbfa}{\mathbf{a}}

\newcommand{\mbfy}{\mathbf{y}}
\newcommand{\mbfc}{\mathbf{c}}

\newcommand{\mbfd}{\mathbf{d}}
\newcommand{\mbff}{\mathbf{f}}

\usepackage{letltxmacro}
\LetLtxMacro\oldttfamily\ttfamily
\DeclareRobustCommand{\ttfamily}{\oldttfamily\csname ttsize\endcsname}
\newcommand{\setttsize}[1]{\def\ttsize{#1}}%
\setttsize{\small}

\title{Knowledge engineering mixed-integer linear programming: constraint typology}

\author{
  Vicky Mak-Hau \\
  School of Information Technology\\
  Deakin University\\
  Waurn Ponds, VIC 3216 Australia \\
  \texttt{vicky.mak@deakin.edu.au} \\
   \And
  John Yearwood \\
  School of Information Technology\\
  Deakin University\\
  Waurn Ponds, VIC 3216 Australia \\
  \texttt{john.yearwood@deakin.edu.au} \\
   \And
  William Moran \\
  Electrical and Electronic Engineering\\
  The University of Melbourne\\
  Parkville, VIC 3010 Australia \\
  \texttt{wmoran@unimelb.edu.au} \\
}

\begin{document}
\maketitle
\begin{abstract}
In this paper, we investigate the constraint typology of mixed-integer linear programming (MILP) formulations. MILP is a commonly used mathematical programming technique for modelling and solving real-life scheduling, routing, planning, resource allocation, timetabling optimization problems, providing optimized business solutions for industry sectors such as: manufacturing, agriculture, defence, healthcare, medicine, energy, finance, and transportation. Despite the numerous real-life Combinatorial Optimization Problems found and solved, and millions yet to be discovered and formulated, the number of types of constraints (the building blocks of a MILP) is relatively much smaller. In the search of a suitable machine readable knowledge representation for MILPs, we propose an optimization modelling tree built based upon an MILP ontology that can be used as a guidance for automated systems to elicit an MILP model from end-users on their combinatorial business optimization problems. Our ultimate aim is to develop a machine-readable knowledge representation for MILP that allows us to map from an end-user's natural language description of the business optimization problem to an MILP formal specification.

\end{abstract}


\section{Introduction} 

Combinatorial Optimization Problems (COPs) arise in many real-life applications such as scheduling \cite{Mak2021,Velez2013}, planning \cite{Akartunali2015,Kruijff2018}, resource allocation \cite{Lalbakhsh2018,Mak2017}, routing \cite{Mak2007,Seixas2013}, and time-tabling \cite{Ghoniem2016,Zhou2020}. See, for example,  \cite{chen2010,Gorman2020,Williams2013}, for more examples of mathematical programming applied in real-life business COPs where millions or even billions of dollars were saved. 

There are several commonly-employed solution approaches for COPs. The two main branches are exact methods and heuristic methods. Mathematical Programming is an exact method that can provide proven optimal solutions, and even when it fails to produce an optimal solution within a predetermined time and memory limit, it can still provide a proven optimality gap. Heuristic approaches (such as trial-and-error, simulation, learning, meta-heuristic or custom-made problem-specific heuristic) on the other hand, do not have a solution guarantee. With meta-heuristics or learning methods, with parameters properly tuned, they may be able to provide reasonably good quality solutions within a much shorter time. Exact algorithms will always be preferred in applications where a proven optimal solution matter. For instance, in Kidney Exchange Optimization, an increment of one unit in the objective function means one more kidney transplant can be carried out, and undoubtably will have a significant impact on the health outcome of the patient with a kidney failure. 

When solving a COP, mathematical Programming-based exact algorithms essentially implement an exhaustive tree search with smart pruning strategies. In the case of Integer Programming (IP)-family of methods, the theoretical basis is algebra, whereas in the case of Constraint Programming (CP), the theoretical basis is logical inferences. CP and IP each has their strengths and weaknesses. IP-family of methods include Pure Integer Programming where all decision variables are integers, Binary Integer Programming where all decision variables are binary, and Mixed-integer Linear Programming (MILP) where some decision variables are continuous, and the rest are binary or general integers. MILP can also model some nonlinear terms (e.g., quadratic, bilinear, and piecewise linear terms), and therefore MILP is a very practical technique in modelling and solving real-life COPs. This is evidenced by the fact that in the history of Franz Edelman Awards, 20\% of the finalists applied IP-family of methods \cite{Gorman2020}, and that the top two algorithms in the 2$^{nd}$ Nurse Rostering Competition are MILP-based methods \cite{Ceschia2019}. 

The formal mathematical specification of a MILP problem is given as follows: 
$\{\min, \ \max\}$    
$\{ \mbfc \cdot \mbfx  + \mbfd \cdot \mbfy \ : \ A\mbfx + B\mbfy \leq \mbff, \ \mbfx \in \mathbb{Z}^n_+, \ \mbfy \in \mathbb{Q}^p_+\}$, 
with $\mbfx=(x_1,\ldots,x_n)$ and $\mbfy=(y_1,\ldots,y_p)$ the decision variables; $\mbfc$ and $\mbfd$ the cost coefficients, for $\mbfc$ a $n$-vector of $\mathbb{Q}$, $\mbfd$ a $p$-vector of $\mathbb{Q}$; $A \in \mathbb{Z}_{m \times n}$ and $B \in \mathbb{Q}_{m \times p}$ the constraint coefficient matrices; and $\mbff$ a $m$-vector of $\mathbb{Q}$. For a thorough exposition of MILP, see, for example, \cite{Nemhauser1998,Pochet2006}. 

An MILP comprises four main components: a set of decision variables, an objective function that is a linear combination of the decision variables, a set of constraints (each containing a linear combination of decision variables), and the index sets that enumerate the decision variables and constraints. We proposed an MILP ontology in \cite{Ofoghi2020}, see Figure \ref{fig:ontology} below. 


\begin{figure*}[ht]
\begin{center}
\includegraphics[width=12cm]{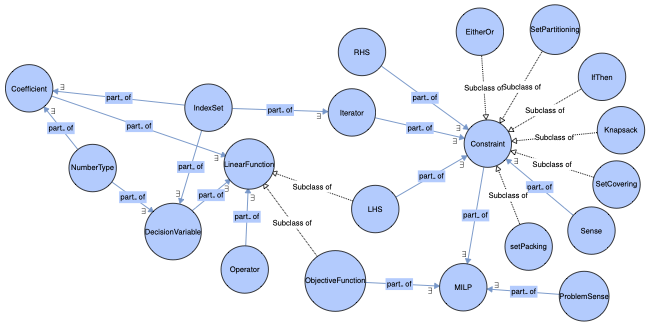}
\caption{The proposed ontology of mixed integer linear programming models for Combinatorial Optimization problems. Note: The visualization is courtesy of WebVOWL~\cite{VOWLpaper}.
}
\label{fig:ontology}
\end{center}
\end{figure*}

\section{Constraint types} 

Here, we ask a fundamental question: is there a finite number of MILP constraint types, and if the answer is yes, how many are there? 
Classic families of COPs such as routing, scheduling, planning have dozens or hundreds of variations from real-life applications. Every COP is different.  Numerous MILPs for real-life COPs found and solved, and many more yet to be discovered and formulated. We are not able to examine every single MILP ever developed in history, however we performed two simple studies for obtaining insights. 1) We examined all constraints used in the production planning problems listed in H. Paul Williams' textbook ``Model building in mathematical programming''. 2) We examined constraints used in a number of publications.

\begin{figure*}[ht]
\begin{center}
\includegraphics[max size={\textwidth}]{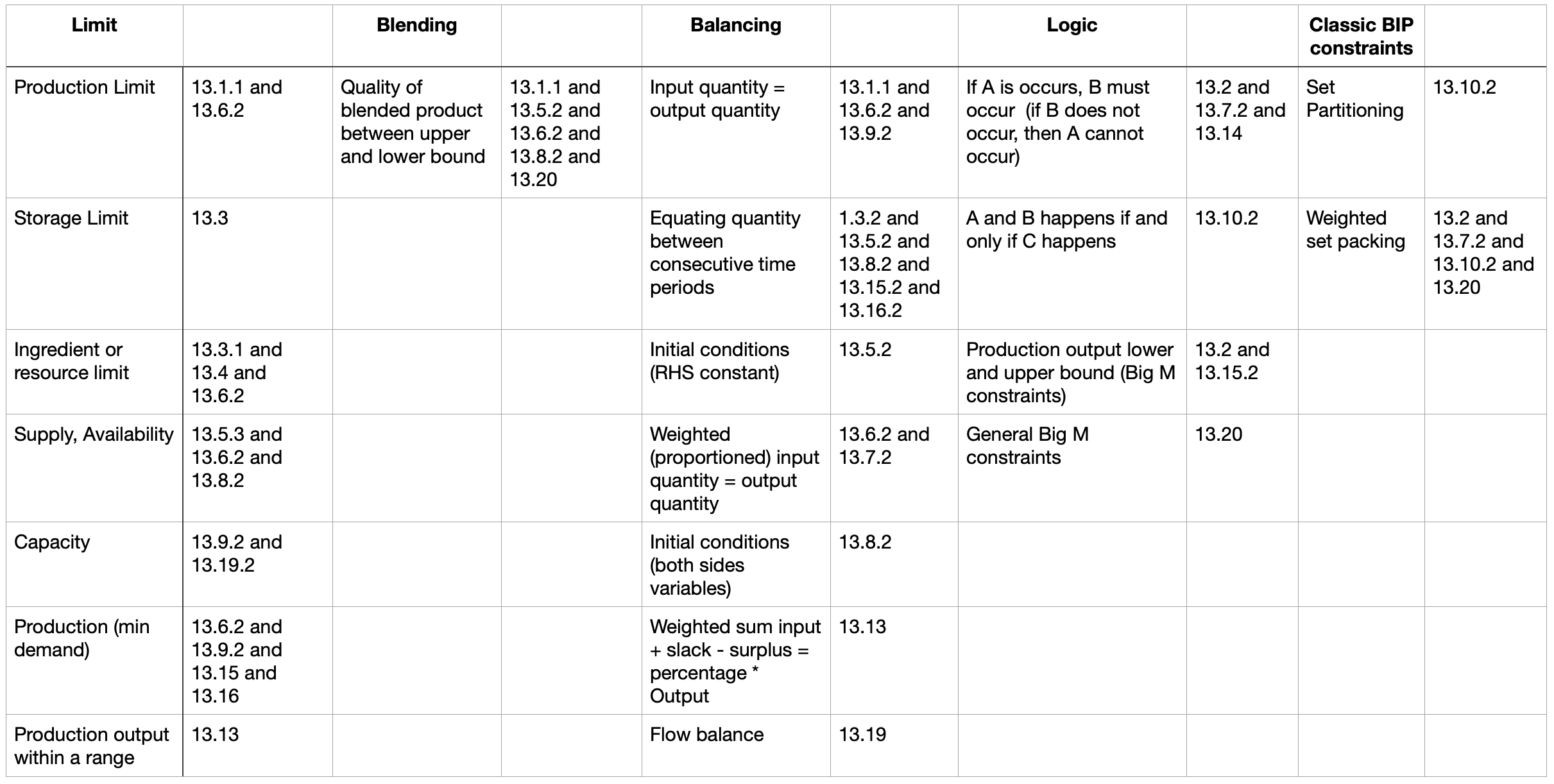}
\caption{A table of constraints and the sections in which they are used in the H Paul Williams book \cite{Williams2013} for production planning problems. 
}
\label{fig:Williams}
\end{center}
\end{figure*}

We considered the production planning examples in H. Paul Williams' book ``Model building in mathematical programming'', and observed that constraints that represent limits (bounds), those that blending from raw materials to products, those that balance two quantities, those that governs logic conditions, and the classic binary integer programming constraints such as set partition, set packing, and set covering and their weighted variations cover all the production planning problems presented therein. In Figure \ref{fig:Williams}, we present a table where we listed the meaning of the constraints used and the section number in the textbook where the examples were discussed. These constraints are reasonably easy to interpret in the sense that the mathematical specification of the constraints is either very close to or can be directly translated from a natural language (NL) description of a n\"{a}ive end-user, (a domain-expert end-user who is not trained to develop MILP models)--we call these explicit constraints.

There are many ways to classify explicit constraints into different types. For instance, if we classify MILP constraint by their mathematical form, they can only be in one of the following forms: $\mbfa \cdot \mbfx \leq b$, $\mbfa \cdot \mbfx = b$, and $\mbfa \cdot \mbfx \geq b$ (here, 
$b\geq 0$). For simplicity, we will write $\mbfa \cdot \mbfx$ as $ax$ for the rest of the paper. 

{\bf Type I: Bound constraints (Demand and Supply)} 
 Resource limit (supply) constraints $ax \leq b$ and demand constraints $ax \geq d$ are very commonly used in MILPs, particularly in production planning-type of problems. For resource limit (supply) constraints, an upper bound on the supply can be fixed (e.g., a Knapsack constraint) or depends on the value of a decision variable. Similar for the lower bound on a  demand that needs to be satisfied. 

{\bf Type II: Balancing constraints} 
Equality constraints $ax = b$ has many variations in its usage: to balance (equate) input and output quantity; to balance the flow or quantities over two consecutive time periods, to set initial conditions, to assign values, and so on. 

{\bf Set packing/partitioning/covering constraints} 
The set packing/partitioning/covering constraints are subtypes of Type I and Type II constraints, typically used for assignment or allocation. They allow us to model the choice of at most/exactly/at least one out of many. The weighted version of the set packing/partitioning/covering constraints allow us to model the choice of $n>1$ out of many. 

{\bf Logic constraints} 
The three main subtypes of {\bf logic constraints} are the {\em Big-M}, {\em If-then}, and {\em Either-or} constraints, each has a number of varieties. (Some of these varieties were discussed in \cite{Ofoghi2020}). 

So, the next question is, what about real-life COPs other than the ones in the \cite{Williams2013} and how do we represent the knowledge in order to enable automatic mapping from business requirements to the mathematical specification (or that of a general purpose modelling language such as OPL or Minizinc)? 

\section{The optimization modelling tree} 
We designed an Optimization Modelling Tree (OMT) and examined a number of COPs in published journal articles to ascertain whether the OMT is adequate in the sense that by traversing through the tree all elements for the MILPs can be found. The focus of this exercise is to evaluate whether the constraint types and subtypes in the OMT are enough to represent these example COPs. 

%
\begin{figure*}[ht]
\begin{center}
\includegraphics[width=16.5cm]{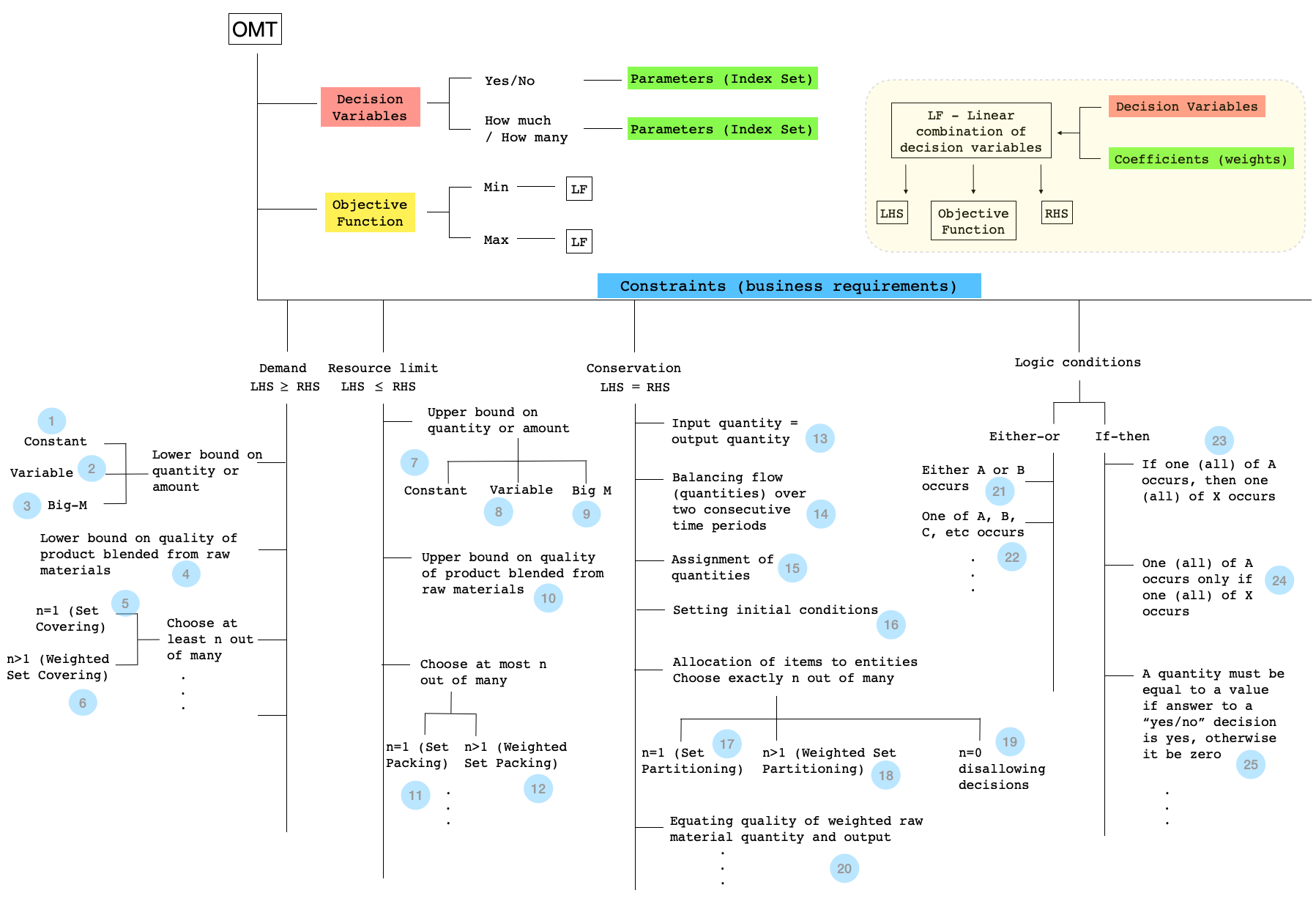}
\caption{The Optimization Modelling Tree (OMT)
}
\label{fig:OMT}
\end{center}
\end{figure*} 
%

{\bf A chemical production scheduling example} 
A chemical production scheduling MILP model is presented in \cite{Velez2013}. The problem considers a given planning horizon, partitioned into a number of time slots. The decisions to be made are whether a unit (a machine or equipment) should start processing a task at a particular time slot, what the batch size should be, and the inventory level of each material at each time slot. A basic MILP is presented with 4 constraints. Constraint Set (1) is a set packing constraint ensuring that each unit will be starting at most one task at a time (Number 11 on the OMT). Constraint Set (2) is a combined logic (Big-M) and upper/lower bound constraint--if a unit starts processing a task at a given time, then the capacity of the batch size must be observed, otherwise, the task will not be processed on this machine at this time (Numbers 3 and 9 on the OMT). Constraint Set (3) presented in the paper should have been two constraints. One is to equate the inventory (storage) of a material at a time slot to the inventory at the previous time slot plus the new production and minus the  consumption (Number 14 on the OMT). The second part of Constraint Set (3) is an upper bound on the storage limit (Number 7 on the OMT). These constraints are reasonably straight forward to describe by an end-user, and all requirements can be found in the constraint types on the OMT. 

{\bf A supply chains production planning example} 
An MILP model for mid-term production planning for high-tech low-volume supply chains is presented in \cite{Kruijff2018}. The decision variables are mostly general integer variables, and the six constraint sets are as follows. Constraint Set (1) are to balance two quantities, in specific, quantities between two consecutive time slots (Number 12 on the OMT). Constraint Sets (2) and (5) are equality constraints for assigning quantities (Number 13 on the OMT). Constraint Sets (3) and (6) are variable upper bounding and lower bounding (Numbers 2 and 8 on the OMT) whereas Constraint set (4) has the logic condition that the upper and lower bounds on decision variables for quantities apply only when the associated binary decision variables is non-zero (Numbers 3 and 9 on the OMT). 

{\bf A university course timetabling problem example} 
A university course timetabling problem was modelled as an MILP in \cite{Ghoniem2016}. The decision variables are binary. One set of the decision variables represent yes/no answers to whether a particular section of a course should be assigned to a particular professor in a particular time slot. Translating them from NL to formal specifications should be reasonably straight forward.  Constraint sets (2) and (3) are Set Partitioning constraints (for choice of exactly one out of many, Number 17 on the OMT), Constraint Sets (4) to (9) are Set Packing constraints (for choice of at most one out of many, Number 11 on the OMT)). Constraint sets (11)--(13), and (15) are general if-then constraints regulating if $X$ occurs, then both of $Y$ and $Z$ must occur. The constraints are in the form of $2X \leq Y + Z$ (Number 24 on the OMT), although $X \leq Y$ and $X \leq Z$ are better constraints to use.  This brings an important aspect for knowledge engineering MILPs: multiple feasible MILP constraints exist for the same requirement, some are strong for computational use than others. The OMT in its current state has some limitations, as we can see from the next example. 

{\bf A multitrip vehicle routing problem with time windows example}. The COP described in \cite{Seixas2013} is a routing-type problem. An end-user not trained with MILP knowledge does not normally describe that a yes/no decision is associated with each pair of locations (e.g., $i$ and $j$ with a yes answer indicating Location $j$ must be visiting immediate after Location $i$). However, commonly-used MILPs for routing problems typically use a binary variable for each of these decisions. Once the hurdle in decision variable definition is overcome, the rest of the constraints can be found in the constraint types or subtypes described in the OMT. Constraint Sets (1), (3), (4) are all Set Partitioning Constraints, i.e., to choose exactly one out of many (Number 17 on the OMT). Constraint Set (2) is to set to zero variables that represent impossible decisions (Number 19 on the OMT). Constraint Sets (6) to (8) are to regulate the time of arrival of a vehicle route to visit a customer, and the constraints are {\em if-then} subtype be found in the OMT. The last constraint set (10) ensures is a straight forward upper bounding constraint on total time used, and the bound itself is a variables (Number 2 on the OMT). 
Constraint Set (9) is a special type of demand - capacity constraint commonly used in routing-type of problems. The requirement is not trivial to describe in NL by an end-user but the mathematical constraint itself can be found in the OMT (it is in fact a Set Packing Constraint, Number 11 on the OMT). Constraint Set (5) is a flow-balance constraint (which is covered by the OMT), the mathematical meaning is that if a customer is visited, then there must be a customer that was visited before him/her and one after him/her. An end-user would not describe the requirement like this. We call these implicit constraints.

\section{Summary remarks and future research} 

What the OMT contains, is not just the mathematical specification of the MILP constarints.  Mathematically, $ax \leq b$, $ax = b$, and $ax \geq b$ are enough to cover all MILP constraints. However, the OMT we designed branches by usage, (or, the meaning of the constraints in application). A beginner MILP modeller, for example, can traverse through the tree to elicit business requirements from a non-expert end-user.  We have tested some COP instances and the constraints in the OMT do in fact cover all the ``explicit'' (or, straight-forward) constraints. Even for constraints in our test cases that are not straight forward, i.e., the ``implicit'' constraints, they too are covered by the OMT, though the mapping mechanism is not represented on the OMT. 

We have the same results for the ACs and the SECs of an ATSP, they can appear in the form of Set Partitioning and Set Covering constraints respectively, but the mapping is not explicitly represented in the tree. 

In summary, we hypothesize that the OMT is scalable, however, we are unable to prove it in this paper. 

\subsection{Future research} 

Now, what is required appears to be the compilation of a list of mappings for commonly-used implicit constraints. For example, the knowledge that ``visiting each city exactly once and return to the home city'' is equivalent to ``each entity has one that precedes it and one that succeeds it'' is needed to be represented on this OMT, and consequently the Set Partitioning and Set Covering constraints be identified as the right ones to use. This will be the subject of our next research project: to perform a  survey of literature for commonly-used implicit constraints and the usage they map to, and to represent such a mapping on the OMT in an efficient way.


\end{document}